# Novel Pigeon-inspired 3D Obstacle Detection and Avoidance Maneuver for Multi-UAV Systems

Reza Ahmadvand, Sarah Safura Sharif, Yaser Mike Banad

*Abstract*—Recent advances in multi-agent systems manipulation have demonstrated a rising demand for the implementation of multi-UAV systems in urban areas, which are always subjected to the presence of static and dynamic obstacles. Inspired by the collective behavior of tilapia fish and pigeons, the focus of the presented research is on the introduction of a nature-inspired collision-free formation control for a multi-UAV system, considering the obstacle avoidance maneuvers. The developed framework in this study utilizes a semi-distributed control approach, in which, based on a probabilistic Lloyd's algorithm, a centralized guidance algorithm works for optimal positioning of the UAVs, while a distributed control approach has been used for the intervehicle collision and obstacle avoidance. Further, the presented framework has been extended to the 3D space with a novel definition of 3D maneuvers. Finally, the presented framework has been applied to multi-UAV systems in 2D and 3D scenarios, and the obtained results demonstrated the validity of the presented method in dynamic environments with stationary and moving obstacles

*Index Terms*— Collision Avoidance, Centroidal Voronoi Tessellation, Distributed Control, Formation Control, Multi-Agent System, Obstacle Avoidance.

## I. INTRODUCTION

Swarm intelligence, an emergent property observed in nature, has long served as a source of inspiration for engineering applications, particularly in the development of autonomous control systems [1]. From an engineering perspective, swarm intelligence shows how decentralized systems, composed of numerous simple agents, can achieve complex collective behaviors. In recent years, the application of multi-agent systems (MAS) in both civil and military domains, such as intelligent transportation systems, surveillance, and search-and-rescue operations, has gained significant attention [2]. Often, in real-world applications, MAS are being implemented in dynamic and complex urban environments in the presence of building static obstacles and dynamic obstacles. Nowadays, along with considering a high priority for the accuracy and autonomy of the manipulated systems, the aspect of safety has become increasingly important. In the mentioned environments, ensuring that the agents can operate safely without colliding to the obstacles or the other vehicles is crucial. Recent efforts in cooperative vehicle control have also shown promise in enhancing safety under extreme scenarios using shared control and potential field-based strategies, such as game-theoretic steering control with fuzzy Koopman modeling for collision avoidance [3]. Although there is a huge literature in the traditional centralized control methods for the multi-UAV systems, their reliability are highly depended on the central computer that is a dominant constraint and in the cases with the failure of the central computer the mission will completely fail [2, 4]. Additionally, control strategies using appointed-time prescribed performance functions and sliding-mode-based fuzzy control have been introduced to ensure convergence in fixed time, but they often rely on structured topologies or high control complexity [5]. Thus, the main aim of the presented study is to introduce a semi-distributed method for the collision-free formation control of a multi-UAV system in the simultaneous presence of the buildings, static obstacles and dynamic obstacles utilizing a novel 3D obstacle avoidance maneuver approach which is the extension of the method proposed for 2D maneuvers in [4]. Previous studies on obstacle avoidance have demonstrated that in addition to the path planning and geometric guidance methods which requires a known map of the implementation environment with modeled obstacles for the UAVs to find a collision-free path between two desired points [6, 7], generally, there are several other categories such as potential field function approaches that work based on the definition of some potential field functions [8], and based on the prediction of considered dynamical model for considered system of UAVs in the future time intervals the model predictive based control approaches have been proposed for the decision about the control inputs [9, 10]. Recent advances in the multi-UAV control systems shows the utilization of stand-alone path planning methods and its combinations with other methods like particle swarm optimization (PSO) methods is still a challenging hot topic for research [11, 12]. Furthermore, based on the artificial potential field (APF) concept, some strategies have been introduced for the implementation of multi-UAV systems in an environment with a known map of the obstacles [13, 14, 15, 16]. In [17] the self-organized obstacle avoidance model for pigeon has been statistically investigated and a mathematical model for their obstacle avoidance maneuver is introduced. Further, the pigeon-inspired avoidance model has been utilized for a group of four quadrotors and the obtained results demonstrated the validity of the method [18]. Although there are many introduced methods for the obstacle avoidance in the presence of static obstacles, the obstacle avoidance in the presence of dynamic obstacles is still a challenging and open problem [4]. Therefore, in [4] a novel pigeon-inspired method has been

Reza Ahmadvand, School of Electrical and Computer Engineering, University of Oklahoma, Norman, OK 73019, USA (e-mail: iamrezaahmadvand1@ou.edu).
Sarah Safura Sharif, School of Electrical and Computer Engineering, University of Oklahoma, Norman, OK 73019, USA (s.sh@ou.edu).
Yaser Mike Banad, School of Electrical and Computer Engineering, University of Oklahoma, Norman, OK 73019, USA (e-mail: bana@ou.edu).
*Corresponding author: Yaser Mike Banad*

proposed for the detection and avoidance of dynamic obstacles. However, they solved the formation control and recovery by utilizing of the consensus theory which adds the complexity to the controller design. On the other hand, they solved the obstacle detection and avoidance considering planar maneuvers which sometimes will yield to constraints and getting the UAVs stuck between their neighbor UAV and an external obstacle which can be a serious limitation particularly for the applications in which the UAV flock needs to pass through a pathway between the walls that is inevitable in the urban environments. So far, the problem of 3D obstacle avoidance maneuvers has been investigated for single UAV problems. In [19] utilizing multi-objective spherical vector-based PSO a path planning approach is introduced for a single UAV problem which has demonstrated successful path planning in 3D environment. Moreover, by hybridizing the slime mould with a different updating algorithm, and employing the Pareto optimality, a novel 3D route planning method for single UAV has been introduced [20]. Therefore, inspired by the works on 3D maneuvers, the pigeon-inspired method introduced in [4], and the collective behavior of tilapia fish, a novel approach with 3D maneuvers has been introduced. Proposed approach in this study, considers three main functionalities for the controller. First, formation control, then inter-vehicle collision avoidance, and finally obstacle avoidance in the presence of static and dynamic obstacles. In this approach, the optimal positioning of UAVs in an optimal formation with respect to their sensing range is inspired by the territorial behavior observed in tilapia fish and implemented by probabilistic Lloyd algorithm [21]. This behavior ensures that the UAVs maintain proper spacing relative to each other, akin to how fish in a school exhibit coordinated patterns. Additionally, each UAV has the ability to independently follow the desired points using a local controller, while, they can have the ability of obstacle detection and performing real-time avoidance maneuvers independent from the other UAVs. Thus, we can have a centralized guidance for the formation control combined with a distributed controller for the rest of the control tasks. Overall, the primary contribution in this study is: 1) development a semi-distributed controller for multi-UAV systems; 2) Novel obstacle detection and avoidance considering 3D maneuvers in the presence of dynamic obstacles. 3) improving the scalability of the previously introduced methods in the multi-UAV system consists of 12 UAVs by utilizing 3D maneuvers while in the previous studies the maximum number of agents was 6.

This paper is organized as follows. Section 2 Initiates with multi-UAV dynamic modeling followed by preliminaries of graph theory. Then the controller design and theoretical developments has first been presented. Section 3 presents the simulations carried out in this study for different case studies. Finally, section 4 concludes the paper and gives some directions for the potential applications and future works.

## II. THEORY

This section, presents the dynamical model of the considered multi-UAV system, followed by the required preliminaries related to graph theory, then the controller design has been outlined.

### A. Multi-UAV Dynamical Model

Fig. 1 schematically depicts the multi-UAV system consist of $N$ UAVs. Also, Fig. 1 demonstrates the inertial coordinate frame (attached to the ground) described by $O_I X_I Y_I$ for the modeling of UAVs translational motion, and the body coordinate frames described by $O_b x_b y_b$ located in the UAVs' center of gravity for the modeling of their relative motion with respect to their neighbor vehicles.

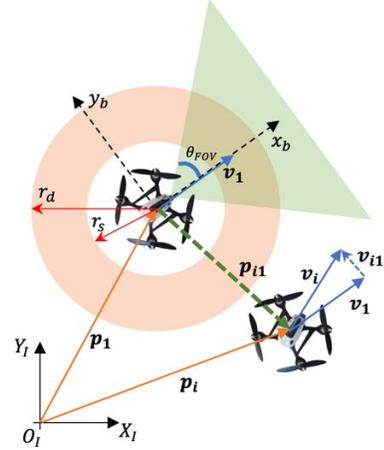

**Fig. 1.** Multi-UAV system dynamic model schematic

Here, $\boldsymbol{p}_i = [p_{xi}, p_{yi}, p_{zi}]$ and $\boldsymbol{v}_i = [v_{xi}, v_{yi}, v_{zi}]$ represent the position and velocity vectors of $i^{th}$ UAV in the inertial coordinate frame that describe the translational subsystem of the dynamical model. $\boldsymbol{p}_{ij} = \boldsymbol{p}_j - \boldsymbol{p}_i$ and $\boldsymbol{v}_{ij} = \boldsymbol{v}_j - \boldsymbol{v}_i$ refer to the relative position and velocity vectors of $i^{th}$ UAV with respect to $j^{th}$ UAV, and the parameters $r_s$ and $r_d$ are safety range and detection range respectively and $\theta_{FOV}$ refers to the field of view of the UAV (with respect to the flight direction). Thus considering $\boldsymbol{\Phi}_i = [\phi_i, \theta_i, \psi_i]$ and $\boldsymbol{\Omega}_i = [p_i, q_i, r_i]$ as the Euler angles tuple and angular velocity vector of $i^{th}$ UAV. The 6 nonlinear dynamical model of a multi-UAV system can be achieved as shown in the following expression [4]:

$$\begin{aligned} \dot{\boldsymbol{p}}_i &= \boldsymbol{v}_i \\ m_i \dot{\boldsymbol{v}}_i &= -m_i g \boldsymbol{k} + \boldsymbol{U}_i^F \\ \dot{\boldsymbol{\Phi}}_i &= \boldsymbol{\Lambda}(\boldsymbol{\Phi}_i) \boldsymbol{\Omega}_i \\ \boldsymbol{J}_i \dot{\boldsymbol{\Omega}}_i &= -\boldsymbol{\Omega}_i \times \boldsymbol{J}_i \boldsymbol{\Omega}_i + \boldsymbol{\tau}_i \end{aligned} \quad (1)$$

In the above equations $\boldsymbol{U}_i^F = [u_{xi}^F, u_{yi}^F, u_{zi}^F]$ are the virtual control inputs to the translational subsystem, $g$ is the gravity constant, $\boldsymbol{k} = [0,0,1]$ is unit vector in the $z$ direction, $\boldsymbol{\tau}_i$ refers to the input vector for the attitude subsystem, $m_i$ and $\boldsymbol{J}_i$ are the mass and inertia matrix of $i^{th}$ UAV. Finally, because the focus of this research is on the high level controller for the translational motion of the UAVs, and also, as it is common in the literature, defining the auxiliary control input $\boldsymbol{u}_i = \boldsymbol{U}_i^F/m_i - g\boldsymbol{k}$ we can use a simplified version of the above mathematical model for our controller design in the next steps [4, 22]:

$$\dot{\boldsymbol{p}}_i = \boldsymbol{v}_i \quad (2)$$

$$\dot{v}_i = u_i \quad (3)$$

In the above, $u_i$ refers to the auxiliary control input for $i^{th}$ UAV.

*B. Graph Theory*

To design our position controller, in this study, the graph theory has been utilized. In the graph theory, a multi-UAV system consist of $n$ homogeneous UAVs will be described using $G(U,E)$ which represents a graph with a set of nodes defined as $U = \{i\}, i = 1,2,...,n$ and a set of $E = \{(i,j)|i,j \in U ; i \neq j\}$ that represents the edges which models the communication between the agents. Here, the set $N_i = \{j \in U \setminus \{i\} | (i,j) \in E\}$ describes the neighborhood for $i^{th}$ node in the considered graph. Thus, if $p_i$, $i = 1, 2, ..., n$ refers to the position vector of $i^{th}$ UAV in the inertial coordinate frame and $r_d$ defines its detection range, the UAVs' neighborhoods can be defined using the following set of nodes in the space:

$$N_i = \left\{ j \in U \setminus \{i\} \, \middle| \, \left\| p_i - p_j \right\| < r_d \right\} \quad (4)$$

*C. Formation Controller Design*

Inspired by the Tilapia fish territorial behavior [23], this section presents the formation controller design for the optimal configuration of UAVs in a barrier planar area $Q$ in the 3D space. Thus, firstly, it is required to find an optimal formation for the UAVs in a predefined area. To this aim the concepts of locational optimization and Voronoi partitions have been leveraged [24, 25, 26]. Therefore, by investigations on the Voronoi configurations in [22], as a matter of fact, the sensing performance of any UAV on an arbitrary point $q$ in its sensing range (belonging to a desired planar barrier area of $Q$) highly depends on the distance $\|q - p_i\|$. Thus, the following multicenter cost function (which can provide a measure of sensing performance expected value provided by all the UAVs at any point $q$ in the space) can be considered for the locational optimization:

$$J(p_1, ..., p_N) = \int \max f(\|q - p_i\|) \phi(q) dq \quad (5)$$

Further, as it has been presented in [22], for an optimal Voronoi partition around $i^{th}$ UAV the following definition can be utilized:

$$V_i = \left\{ q \in Q : \|q - p_i\| \leq \|q - p_j\|, j \neq i \right\} \quad (6)$$

The above definition for the Voronoi partitions, and the cost function defined in Eq. (5) leads to the following differential equation which shows the differentiation of considered cost function with respect to UAVs location in inertial frame:

$$\frac{\partial J_v}{\partial p_i} = M_{v_i}(c_{v_i} - p_i) \quad (7)$$

The above expression demonstrates that the optimal solution for the considered cost function where $\frac{\partial J_v}{\partial p_i} = 0$ is on the locations that satisfy $p_i = c_{v_i}$ which leads to a centroidal Voronoi tessellation (CVT). Note that, In the Eq. (7) the parameter $M_{v_i}$ is the Voronoi center of mass, and $c_{v_i}$ is Voronoi centroid. So far it has been demonstrated that the optimal configuration for the UAVs is the CVT. Therefore, to address this aspect and finding the optimal points, the probabilistic generalized Lloyd's algorithm has been utilized [21]. The pseudo-code of implemented Lloyd's algorithm has been provided in the Algorithm 1.

---

**Algorithm 1** Probabilistic Generalized Lloyd's Algorithm

**Input:**
- Predefined bounded area $Q \subset R$
- Density function $\rho(x)$ defined on $Q$
- Positive integer $N$ (number of UAVs)
- Positive integer $S_{num}$ (number of sampling points per iteration)
- Constants α₁, α₂, β₁, β₂ such that α₁ + α₂ = 1, β₁ + β₂ = 1, α₂ > 0, β₂ > 0

**Output:**
- Final set of points $\{x_i\}N_{i=1}$.

**Algorithm:**

1. Initialization:
- Choose an initial set of $N$ points $\{x_i\}N_{i=1}$ in $Q$.
- Set iteration counters $\{j_i\}N_{i=1} = 1$.

2. Sampling:
- Randomly sample $S_{num}$ points $\{y_r\}Snum_{r=1}$ in $Q$ using a uniform distribution $\rho(x)$ as the probability density function.

3. Point Update:
   For each i = 1, 2, ..., N:
   - Gather all sampled points $y_r$ closest to $x_i$ (forming the set $W_i$, i.e., the Voronoi region of $x_i$).
   - If $W_i$ is empty, do nothing.
   - Otherwise:
      - Compute the average $u_i$ of the points in $W_i$.
      - Update $x_i$:
        $x_i \leftarrow ((α_1 j_i + β_1) / (j_i + 1)) x_i + ((α_2 j_i + β_2) / (j_i + 1)) u_i$
      - Increment $j_i$:
        $j_i \leftarrow j_i + 1$

4. Repeat or Terminate:
- Form the new set of points $\{x_i\}N_{i=1}$.
- If $Q$ is a hypersurface, project $x_i$ onto $Q$
- Check stopping criteria (e.g., convergence or tolerance).
- If criteria are not met, go back to Step 2.

---

Next, to apply the obtained results from the CVT algorithm and have an optimal configuration for UAVs in our considered barrier area $Q$, the following control law needs to be utilized:

$$u_{f_i} = -K_{p_i}(p_i - c_{v_i}) - K_{v_i} v_{i_{rel}} \quad (8)$$

In the above expression, the gain $K_{p_i}$ and $K_{v_i}$ are positive definite matrices, which can be tuned by trial and error or any design method, $v_{i_{rel}}$ is the relative velocity between $i^{th}$ UAV and their desired point on the considered barrier, which is moving with the speed of 1 m/s in the $x$ direction. Fig. 2 compares the obtained result from the simulation of introduced Lloyd's algorithm in Table 1 for 20 seeds (UAVs for this research) with the real-world photograph, which shows Tilapia fish school territorial behavior. The comparison here verifies the inspiration of our method from Tilapia fish school. Moreover, the demonstrated pattern for the seeds (Fig. 5(a)) is the optimal positioning configuration based on their sensing range.

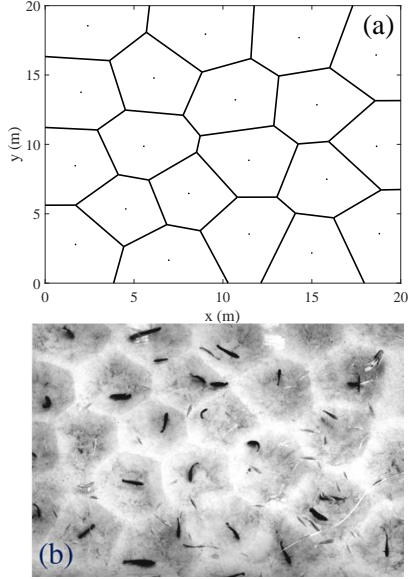

**Fig. 2.** Comparison of the, a) obtained optimal pattern from Lloyd's algorithm in the predefined barrier area with, b) the real-world photograph which shows Tilapia fish school territorial collective behavior [23].

Finally, using the approach proposed in this section any number of UAVs can be optimally deployed in a desired barrier area for either stationary or moving barriers. Thus, the formation controller pushes the UAVs to follow and keep tracking this pattern even when they are moving. (Notice: the considered barrier must be large enough to enclose the UAVs with predefined safety areas around them.)

*D. Collision Avoidance Controller Design*

In the preceding section, the inter-vehicle collision avoidance which is a crucial factor for the multi-UAV systems will be presented. Inspired by the Hook's law and the spring-mass-damper system, considering the relative position and velocities between the vehicles, in [4], the following control law has been introduced:

$$u_{c_{ij}} = -k_{c1}p_c + k_{c2}v_{ij} \qquad (9)$$

where:

$$p_c = \frac{1}{\left(\|p_{ij}\| - r_s\right)^2} \frac{p_{ij}}{\|p_{ij}\|} \qquad (10)$$

In the above equations, $k_{c1}$ and $k_{c2}$ are diagonal positive definite matrices that can be tuned by trial and error. The pseudo-code for the collision avoidance algorithm has been provided in Algorithm 2.

---
**Algorithm 2** Inter-vehicle collision avoidance
---
**Input:**
 NOA: Number of UAVs
 $p_i$: Position vector of $i^{th}$ UAV ($i = 1\ to\ N$)
 $v_i$: Velocity vector of $i^{th}$ UAV
 $k_{c1}$: Diagonal positive-definite gain matrix
 $k_{c2}$: Diagonal positive-definite gain matrix
 $r_s$: safety range
 $N_i$: Set of neighboring UAVs for UAV i
**Output:**
 $u_{ci} \leftarrow$ Collision avoidance control input for $i^{th}$ UAV
**Algorithm:**
 FOR each UAV i in 1 to NOA do:
  If i ≠ j then:    # Ignore self
   $p_{ij} = p_j - p_i$   # Relative position
   If $\|p_{ij}\| < r_s$ then:
    Activate the collision avoidance (Neighbor detected)
    $p_c = 1/\left(\|p_{ij}\| - r_s\right)^2 (p_{ij}/\|p_{ij}\|)$
    $v_{ij} = v_j - v_i$    # Relative velocity
    $u_{c_{ij}} = -k_{c1}p_c + k_{c2}v_{ij}$    # Control input
   ELSE:
    $u_{c_{ij}} = 0$   # Set to zero
   END IF
  ELSE
   $u_{c_{ij}} = 0$   # Set to zero
  END IF
 END FOR
---

*E. Obstacle Avoidance Controller Design*

The focus of this section is on the Pigeon-inspired obstacle avoidance method which has previously been proposed for 2D maneuvers. Thus, the present section, firstly, reviews the previously introduced obstacle avoidance controller design introduced for planar 2D maneuvers in the presence of the moving obstacles while the obstacles are constrained to just move in the perpendicular direction to the UAV flock movement direction [4]. Further, by reformulating the equations, the method will be extended into the 3D space for nonplanar maneuvers which will give the UAVs an extra degree of freedom in obstacle avoidance. Generally, throughout this study, the obstacle avoidance maneuver consists of two parts: First, the obstacle detection, then, the maneuver execution using velocity adjustment. So, they have been outlined separately.

*Planar Obstacle Detection (2D)*

Inspired by the pigeon sector-like field of view (FOV), Fig. 3 schematically shows the obstacle detection of UAVs with limited FOV.

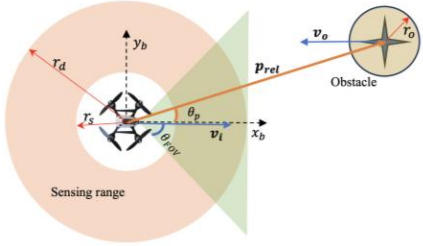

**Fig. 3.** Schematic representation of obstacle detection, illustrating the sensing range and FOV.

Considering limited FOV for the pigeons in previous works [4], has led to the following expressions for the obstacle detection:

$$\|\mathbf{p}_i - \mathbf{o}_k\| \le r_d + o_{rk} \tag{11}$$

$$\left|\operatorname{atan}\left(\frac{o_{yk} - p_{yi}}{o_{xk} - p_{xi}}\right) - d_i^{fly}\right| < \theta_{FOV}$$
$$\text{or } \left|\operatorname{atan}\left(\frac{o_{zk} - p_{zi}}{o_{xk} - p_{xi}}\right)\right| \le \theta_{FOV} \tag{12}$$

In the above, the $\mathbf{o}_k = [o_{kx}, o_{ky}, o_{kz}]$ is the position vector of $k^{th}$ obstacle in the UAV body coordinate frame. Here, Eq. (11) checks if the obstacle is in the sensing range or not, meanwhile Eq. (12) determines the presence of the obstacle in the FOV. In other words, the detection rule here says that if the distance between $i^{th}$ UAV and $k^{th}$ obstacle is in the detection range and the angle between the flight direction and $\mathbf{p}_{rel}$ is less than considered FOV, the obstacle will be detected.

*Planar Velocity Adjustment (2D)*

Considering the location of the obstacle and current velocity of the UAV in the body coordinate frame and defining a potential function introduced in Eq. (13) has led to the velocity adjustment method introduced here which has been previously proposed for the obstacle avoidance maneuvers [4]:

$$U_p(\mathbf{p}_i, \mathbf{o}_k) = \begin{cases} \frac{1}{2}(\|k_v(\mathbf{p}_i - \mathbf{o}_k)\| - r_a)^2, & detected \\ 0, & otherwise \end{cases} \tag{13}$$

In the above, $k_v$ is a positive definite diagonal matrix, where $k_v = diag\{k_x^o, k_y^o\}$, and $r_a = r_d + r_{ok}$. Differentiating the presented potential function with respect to $p_i$ (position vector of $i^{th}$ UAV) leads to the following expression:

$$\nabla U_p = \begin{cases} (\|k_v(\mathbf{p}_i - \mathbf{o}_k)\| - r_a)\nabla(\|\mathbf{p}_i - \mathbf{o}_k\|), & detected \\ 0, & otherwise \end{cases} \tag{14}$$

In the above equation, $\nabla$ refers to the gradient operator, and $\nabla(\|\mathbf{p}_i - \mathbf{o}_k\|) = (\mathbf{p}_i - \mathbf{o}_k)/\|\mathbf{p}_i - \mathbf{o}_k\|$. Now, considering the above obtained expression will yield to a situation in which our UAVs will get stuck with a velocity of 0 m/s which can be considered as a local minima convergence when the force in the forward direction of the UAV is equal to the force of the evasive obstacle. Thus, to avoid such conditions, a rotational potential function has been considered which will be designed using the following procedure:

$$U_r(\mathbf{p}_i, \mathbf{o}_k) = \begin{cases} \frac{k_r}{2}(\|\mathbf{T}_r(\mathbf{p}_i - \mathbf{o}_k)\|)^2, & detected \\ 0, & otherwise \end{cases} \tag{15}$$

Differentiation with respect to $\mathbf{p}_i$ yields to the following expression:

$$\nabla U_r = \begin{cases} k_r \mathbf{T}_r(\|(\mathbf{p}_i - \mathbf{o}_k)\|)\nabla(\|\mathbf{p}_i - \mathbf{o}_k\|), & detected \\ 0, & otherwise \end{cases} \tag{16}$$

Where, $k_r$ refers to a positive coefficient which can be tuned by trial and error, and $\mathbf{T}_r$ is rotation matrix in the $xy$ plane defined as below:

$$\mathbf{T}_r = \begin{bmatrix} \cos\alpha & -\sin\alpha \\ \sin\alpha & \cos\alpha \end{bmatrix} \tag{17}$$

Where:

$$\alpha = \begin{cases} \frac{\pi}{2}\frac{r_d - \|\mathbf{p}_{rel_{ik}}\|}{r_d - r_s}, & r_s < \|\mathbf{p}_{rel_{ik}}\| < r_d \\ 0, & otherwise \end{cases} \tag{18}$$

In the above expression, $\mathbf{p}_{rel_{ik}}$ refers to the relative position vector between $i^{th}$ UAV and $k^{th}$ obstacle in the space. Finally, the obstacle avoidance control law will be achieved as it has been expressed below [4]:

$$\mathbf{u}_{oi} = -k_{o1}\sum_{k=1}^{K}\nabla U_p - \sum_{k=1}^{K}\nabla U_r - k_{o2}\mathbf{v}_i \tag{19}$$

Where, $k_{o2}$ is a diagonal positive definite matrix which can be tuned using trial and error. For the proof of the stability the readers can refer to [4]. As it has been Using presented control law in Eq. (19), the UAVs can perform planar obstacle avoidance in their motion surface. Although the reviewed method here, has exhibited successful utilizations in the previous works, it has a limitation which is one of the concerns of the presented study. In some environmental situations like what that has been schematically shown in Fig. 4(a), in which the UAV1 is encountering with the obstacle, there are two possible pathways for it to pass through the obstacle, but the pathway A is temporarily blocked by the UAV2, and the pathway B is blocked by the building wall. thus, sometimes in the presence of the neighbor UAVs, obstacles and buildings it can be challenging for the UAVs (like UAV1 here) to perform their obstacle avoidance maneuver simultaneously with collision avoidance and they might get stuck for a while. Such situations which is inevitable in real-world application can be considered as unpredicted environmental singularities that can reduce the maneuverability of the flock because in such situations the flock will experience a high entropy deformation from the optimal positioning of UAVs. On the other hand, it can slow down the flock movement or causes latency information recovery. The aim of the presented study is to address the mentioned limitation and give more degrees of freedom to the UAVs to perform non-planar 3D obstacle avoidance maneuvers in the space like what it is being demonstrated in Fig. 4(b). In this approach, meanwhile the pathways A and B are blocked, the UAV1 can go outside of the flock-motion plane towards the $z$ direction and make a new pathway like C which is always available. Leveraging these types of maneuvers will solve the mentioned limitations while reduce the overall control effort of the flock by changing the global decision of the flock for the formation change to pass these situations into a local individual decision making and pathway change by a single UAV in the flock. Thus, this approach will also give us more computational and energy efficiency in real-world applications.

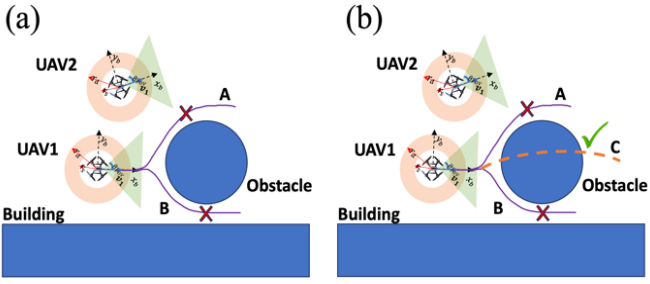

**Fig. 4.** Schematic representation of non-planar obstacle avoidance

To develop the mentioned 3D maneuvers in the next part, rotational matrices, the concept of 3D rotations in the space [27], and the inner product from the linear algebra have been implemented.

*Nonplanar Obstacle Detection (3D)*
As it has been depicted in Fig. 5, unlike the assumption of previously introduced method, here the sensing rang and safety range is assumed to be spherical areas around the UAVs, and the FOV is assumed to be conical region around the flight direction.

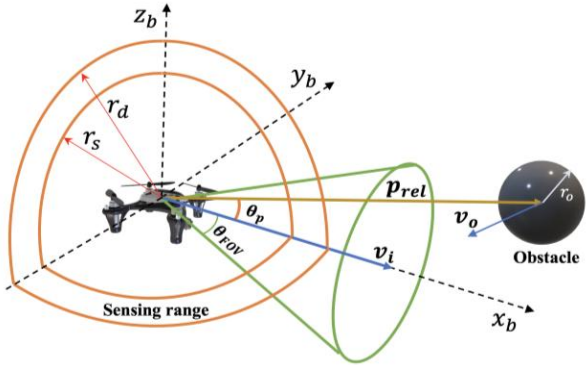

**Fig. 5.** Schematic representation of obstacle detection strategy, illustrating the sensing range and FOV in 3D space

Here, in the Fig. 5, the parameter $\theta_p$ is the angle between the obstacle position vector $\boldsymbol{p}_{rel}$ and the velocity vector $\boldsymbol{v}_i$. Considering the mentioned assumptions, and the implementation of vectors inner product properties in the space led to the following set of conditions for the obstacle detection:

$$||\boldsymbol{p}_i - \boldsymbol{o}_k|| \leq r_d + o_{rk} \quad (20)$$

$$\left|\left(\cos^{-1}\left(\frac{\boldsymbol{p}_{rel} \cdot \boldsymbol{v}_i}{||\boldsymbol{p}_{rel}||||\boldsymbol{v}_i||}\right)\right)\right| \leq \theta_{FOV} \quad (21)$$

In this approach, the same as previous method, the Eq. (20) checks if the obstacle is in the sensing range of the UAV or not. And then using the inner product of the position vector of obstacle and UAV velocity vector, the relative angle between these two vectors can be calculated and if the obtained angle was less than $\theta_{FOV}$ it means that the obstacle is in the UAV field of view. Thus, if both the above conditions are simultaneously met it means that the obstacle is in the sensing range and in the UAV's FOV, so it is detected, and the obstacle avoidance maneuver needs to be executed.

*Nonplanar Velocity Adjustment (3D)*
In the previously presented method for planar maneuvers, the rotation matrix utilized in the proposed potential function in Eq. (16) corresponds to a planar rotation in the $O_b x_b y_b$ plane around the $z$ direction of the body coordinate frame. Thus, in this study, for the extension of the presented pigeon-inspired maneuver into the 3D space, the rotational potential function presented in Eq. (16) needs to be reformulated for a three-dimensional rotation matrix when adjusting the velocity vector. Thus, we can consider a new rotational potential function as below:

$$U_r(\boldsymbol{p}_i, \boldsymbol{o}_k) = \begin{cases} \frac{k_r}{2}\left(||\boldsymbol{T}_{rx}\boldsymbol{T}_{ry}\boldsymbol{T}_{rz}(\boldsymbol{p}_i - \boldsymbol{o}_k)||\right)^2, & detected \\ 0, & otherwise \end{cases} \quad (22)$$

Where:

$$\boldsymbol{T}_{rx} = \begin{bmatrix} 1 & 0 & 0 \\ 0 & 1 & 0 \\ 0 & 0 & 1 \end{bmatrix} \quad (23)$$

$$\boldsymbol{T}_{ry} = \begin{bmatrix} \cos\alpha & 0 & \sin\alpha \\ 0 & 1 & 0 \\ -\sin\alpha & 0 & \cos\alpha \end{bmatrix} \quad (24)$$

$$\boldsymbol{T}_{rz} = \begin{bmatrix} \cos\alpha & -\sin\alpha & 0 \\ \sin\alpha & \cos\alpha & 0 \\ 0 & 0 & 1 \end{bmatrix} \quad (25)$$

In the above expression, $\boldsymbol{T}_{rx}$, $\boldsymbol{T}_{ry}$ and $\boldsymbol{T}_{rz}$ are rotation matrix for the rotations around $x_b$, $y_b$ and $z_b$ respectively. Note that, the presented potential function here now has the ability to adjust the velocity vector direction for the directions outside the motion-plane, but here we considered an identity matrix for the rotation around $x_b$ because we didn't want our UAVs to have such rotations to maintain the attitude stability. Next, differentiation with respect to $\boldsymbol{p}_i$ yields to the following:

$$\nabla U_r = \begin{cases} k_r \boldsymbol{T}_{rx}\boldsymbol{T}_{ry}\boldsymbol{T}_{rz}(||(\boldsymbol{p}_i - \boldsymbol{o}_k)||)\nabla(||\boldsymbol{p}_i - \boldsymbol{o}_k||), \\ 0, \end{cases} \quad (26)$$

Now, utilizing the above introduced 3D rotational potential function and its derivative, in the obstacle avoidance rule presented in Eq. (19) can lead to a velocity adjustment rule for the 3D obstacle avoidance approach. It is notable that, since we neither change the control law nor change the main assumptions, the stability conditions presented in [4] will not change so the method obeys the proof of stability provided in [4]. The pseudo-code for the presented obstacle detection and avoidance in this part has been provided in Algorithm 3.

---

**Algorithm 3** Non-planar obstacle detection and avoidance

**Input:**
NOA: Number of UAVs
$\boldsymbol{p}_i$: Position vector of $i^{th}$ UAV ($i = 1$ to $N$)
$\boldsymbol{v}_i$: Velocity vector of $i^{th}$ UAV
$k_{o1}$: Diagonal positive-definite gain matrix
$k_{o2}$: Diagonal positive-definite gain matrix
$k_r$: Diagonal positive-definite gain matrix
$r_s$: safety range
$N_i$: Set of neighboring UAVs for UAV i
$\boldsymbol{o}_k$: Position of $k^{th}$ obstacle
$r_d$: Detection range
$r_s$: Safety range
$\theta_{FOV}$: Half-angle of the conical FOV
obs_radius: Radius of obstacles sphere
distance_to_obstacle = Distance to obstacle

**Output:**
$\boldsymbol{u}_{oi} \leftarrow$ Collision avoidance control input for $i^{th}$ UAV

**Algorithm:**
FOR each obstacle $k$ ($k = 1,2,...,K$) DO:
  Compute relative position vector and distance to obstacle:
  $\boldsymbol{p}_{rel} = \boldsymbol{p}_i - \boldsymbol{o}_k$
  distance_to_obstacle = $||\boldsymbol{p}_{rel}||$
  Compute $\theta_p$ (the angle between $\boldsymbol{p}_{rel}$ and $\boldsymbol{v}_i$):
  $\theta_p = |(\cos^{\wedge}(-1)((\boldsymbol{p}_{rel}.\boldsymbol{v}_i)/(||\boldsymbol{p}_{rel}||||\boldsymbol{v}_i||)))|$
  IF distance_to_obstacle < (obs_radius + $r_d$) AND $\theta_p < \theta_{FOV}$ THEN:
    Activate the obstacle avoidance (obstacle detected)
    Gradient of translational potential field Eq. (13):
    $\nabla U_p = (||k_v(\boldsymbol{p}_i - \boldsymbol{o}_k)|| - r_a)\nabla(||\boldsymbol{p}_i - \boldsymbol{o}_k||)$
    Rotation angle:
    $\alpha = \pi/2 \, (r_d - ||\boldsymbol{p}_{rel_{ik}}||)/(r_d - r_s)$
    Gradient of rotational potential field Eq. (15):
    $\nabla U_r = k_r \boldsymbol{T}_{rx}\boldsymbol{T}_{ry}\boldsymbol{T}_{rz}(||(\boldsymbol{p}_i - \boldsymbol{o}_k)||)\nabla(||\boldsymbol{p}_i - \boldsymbol{o}_k||)$
    Obstacle avoidance control input Eq. (19):
    $\boldsymbol{u}_{oi} = -k_{o1}\sum_{k=1}^{K}\nabla U_p - \sum_{k=1}^{K}\nabla U_r - k_{o2}\boldsymbol{v}_i$
  ELSE
    $\boldsymbol{u}_{oi} = \boldsymbol{0}$    # Set to zero
  END IF
END FOR

---

*F. Overall Control Logic for Multi-UAV System*

Based on the previously introduced controller designs for different aspects of our problem, and utilizing the super position principle the overall control law for each UAV obeys the following expression:

$$\boldsymbol{u}_i = \boldsymbol{u}_{fi} + \boldsymbol{u}_{ci} + \boldsymbol{u}_{oi} \quad (27)$$

Remarkably, in the above presented expression, the first term in our control input is responsible for formation control and forces the UAVs to follow their own desired points obtained from the Lloyd's method while they are moving with a non-static barrier area, and the second and third terms are responsible for inter-vehicle collision avoidance and obstacle avoidance respectively. Also, it is notable that, the control logic considered in our approach has been demonstrated in the Fig. 6. The figure shows that in our approach the controllers for the inter-vehicle collision and obstacle avoidance are not activated all the time they will just be activated depending on the detection of neighbor UAVs or obstacles.

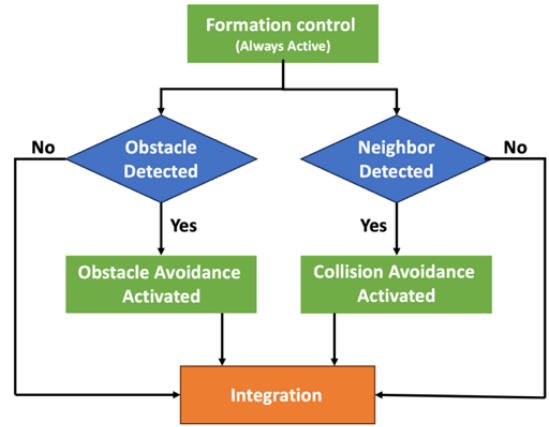

**Fig. 6.** Flowchart of control logic considered in the proposed control approach

## III. NUMERICAL SIMULATION

To evaluate the validity of the proposed method in this study, three case studies have been investigated. Firstly, to assess the performance of the proposed method, it has been applied to the scenario of passing through a narrow pathway (analogy to the pathways between two buildings) considering static and dynamic obstacles. Finally, the proposed method has been applied to the same problem in the 3D space utilizing the 3D maneuvers.

*A. Case-Study 1: Collision-free Formation Control with Obstacle Avoidance using Planar Maneuvers*

To evaluate the validity of the presented controller in a complex situation, in this section it has been applied to the scenario of collision-free formation control of a UAV flock consist of 8 UAVs considering planar obstacle avoidance maneuvers in the simultaneous presence of buildings, static obstacles and a dynamic obstacle the considered scenario here is like that, first the controller deploys the UAV flock into a predefined planar barrier area with 5 meters of altitude in the inertial coordinate frame while it is using the obtained results from the utilized Lloyd's algorithm for the optimal configuration of the UAVs. Then the flock move towards the x-direction while encountering with mentioned buildings and obstacles. Simulation in this section has been performed using the numerical values provided in TABLE I for a time duration of 145 seconds with a time step of 0.1 second. It is notable that the initial position of UAVs have been randomly sampled from the set $\{(x_{0_i}, y_{0_i}, 0) | 0 < x_i < 1, 0 < y_i < 1 : i = 1,2,...,8\}$ using a zero mean Gaussian distribution with the covariance of 1 considering them stationary at $t = 0$s. Moreover, the static obstacles have been located at the points demonstrated in the figures, and the dynamic obstacle is moving with a constant velocity vector of $v_{obs4} = [0.1, 0.025, 0]^T$ m/s. Also, the obstacle 4 (obs4) starts moving after is in stationary condition before $t = 42$s.

TABLE I
NUMERICAL VALUES UTILIZED IN SIMULATIONS

| Parameter | Value |
|---|---|
| $r_d (m)$ | 2 |
| $r_s (m)$ | 1 |
| $K_p$ | $diag([3,3,3])$ |
| $K_v$ | $diag([5,5,5])$ |
| $k_v$ | $diag([0.1,0.5,0.1])$ |
| $k_r$ | 0.5 |
| $k_{o1}$ | 5 |
| $k_{o2}$ | 1 |
| $v_{obs4}$ (m/s) | $[0.2,0.05,0]$ |
| $r_{ok}$ (m/s) | 1 |
| $\theta_{FOV}$ (deg) | 60 |

Fig. 7 demonstrates the obtained result for the utilization of the Lloyd's algorithm for the initial optimal configuration of the UAV flock. It also demonstrates that in our implementation for the formation controller; after finding the optimal desired positioning, the controller allocates each point to the nearest UAV based on their initial position, and in the rest of the scenario each UAV will follow its own desired allocated position while the barrier is moving.

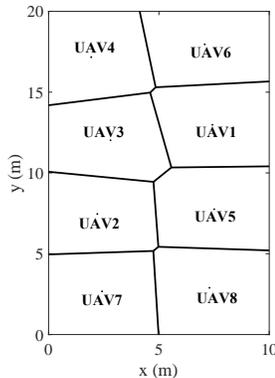

**Fig. 7.** optimal positioning configuration obtained from the Lloyd's algorithm for the formation control

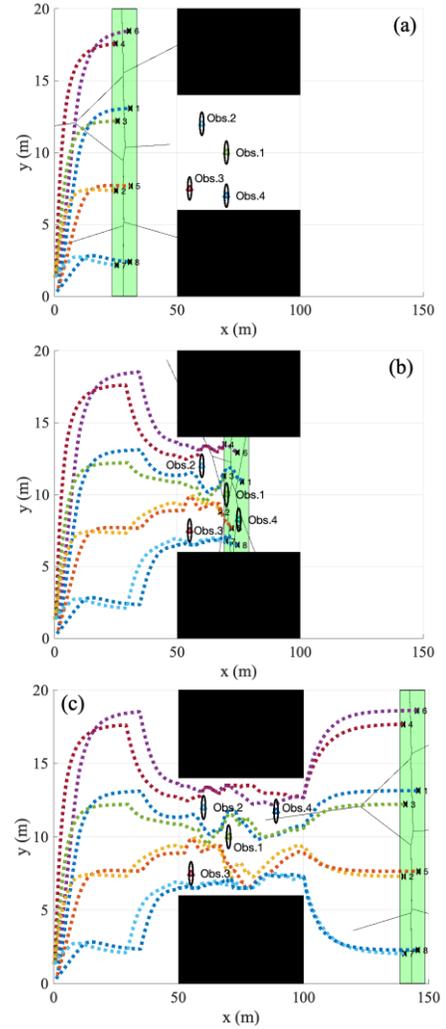

**Fig. 8.** Obtained trajectories for collision-free control scenario with planar obstacle avoidance maneuver

Fig. 8 shows the obtained trajectories for the whole simulation of the scenario in different time frames. Fig. 8(a) shows how satisfactorily the formation controller has deployed the UAV flock in their desired locations (demonstrated in Fig. 7), in the predefined barrier area which has been highlighted here with a green rectangle. Fig. 8(b) shows, how in the presence of the buildings, the formation controller has adopted a new compact formation (by scaling the initial formation while the inter-vehicle collision avoidance is activated) for entering to the pathway between the buildings. Also, the Fig. 8(b) illustrates the UAVs' ability to effectively avoid collisions with obstacles without requiring data exchange between vehicles. This capability stems from the design of both inter-vehicle collision and obstacle avoidance controllers, which operate locally and independently for each UAV, enabling fully distributed control for autonomous decision making. Moreover, the figure demonstrates the formation adaptation of the UAV flock as they approach the narrow pathway between buildings where it is crowded with obstacles. At this critical juncture, each UAV must maintain its situational awareness of neighboring vehicles to prevent inter-vehicle collisions while simultaneously executing obstacle avoidance maneuvers and avoiding contact with building walls. Fig. 8(c) demonstrates the successful passage of the UAVs from the considered area, and it demonstrates the complete formation recovery process of the UAV flock. Initially, we observe the successful passage of the UAVs past the obs4 in obstacle cluster and their immediate formation adjustment for recovering the initial formation as they navigate through the confined area between buildings confirming the system's ability to maintain formation integrity despite environmental challenges. Fig. 9 shows the obtained relative distance between the UAV4 with all other UAVs in the flock meanwhile the horizontal dashed line in the figure shows the boundary of the safety range for the UAV4. Here, the solid lines plotted in the figure reveals the variation of relative distance $|d_{ij}|$ where, $\{(i,j)|i \neq j, j = 1,2,...,8\}$ between the UAVs and the target UAV4. It is clearly evident that all the neighbor UAVs have effectively utilized the inter-vehicle collision avoidance even during the formation change, performing obstacle avoidance maneuvers and formation recovery at the end. Note that, here we just plotted the results for the $i = 4$. (Here, the UAV4 has been selected randomly)

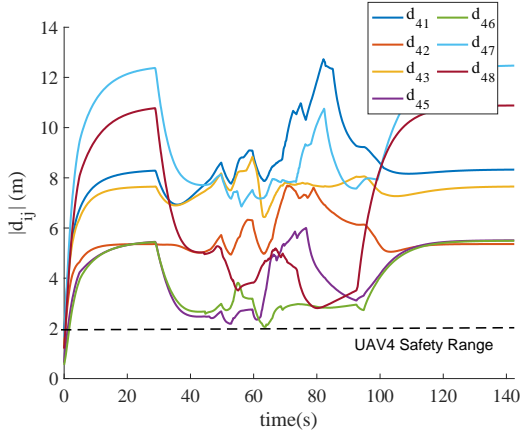

**Fig. 9.** Obtained results for the relative distance with planar maneuvers

Overall, the obtained results from the simulations in this section have demonstrated the effective utilization of the method proposed for the semi-distributed control approach for the safe collision-free formation control missions manipulating UAV flocks in complex environmental situations like urban areas.

*B. Case-Study 2: Collision-free Formation Control with Obstacle Avoidance using 3D Maneuvers*

This section presents the simulation results obtained for the proposed 3D maneuvers applied to a UAV swarm consisting of 12 UAVs in the presence of buildings, static and dynamic obstacles. For the simulations in this section the same scenario as the previous section has been considered in which the UAVs need to be deployed to a predefined barrier area with an optimal configuration while they are moving toward the x-direction, encountering the buildings and obstacles. But here the UAVs are using the 3D approach for detection and avoidance of obstacles. Also, for the simulations conducted in this section the same numerical values as the TABLE I have been utilized. Fig. 10 is demonstrating the obtained trajectories for the whole simulation of the scenario in different time frames.

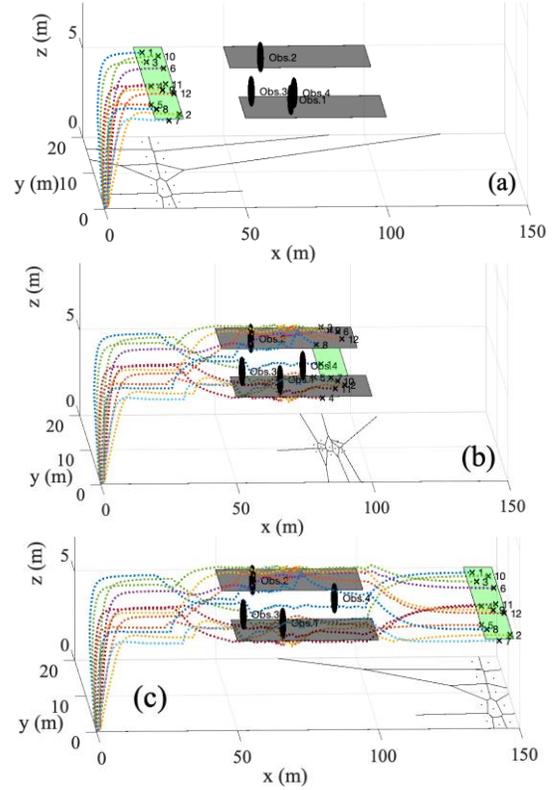

**Fig. 10.** Simulation results for collision-free control scenario with nonplanar 3D obstacle avoidance maneuver

Fig. 10(a) shows the successful deployment of the UAVs in a 3D barrier space utilizing the desired optimal positioning points determined by the Lloyd's algorithm in the space, and the projections of the UAVs on the ground have been demonstrated. Fig. 10(b) depicts how the flock has performed the formation change into a smaller area in the space by scaling the initial formation to be able to pass through the narrow space between the buildings while the UAVs are aware of their neighbors for utilizing the inter-vehicle collision avoidance. Also, the figure demonstrates how satisfactorily the UAVs have performed the obstacle detection and avoidance using the designed 3D approach for the detection and maneuvers in the space while they are utilizing the inter-vehicle collision avoidance simultaneously with keeping themselves away enough from the building's wall boundaries. Fig. 10(c) illustrates the successful formation recovery of the flock after passing the buildings and obstacles. It is noteworthy that in this section the building walls have not been plotted in the figures for the better visibility of the trajectories and performed maneuvers. Thus, just the ceilings have been depicted here.

10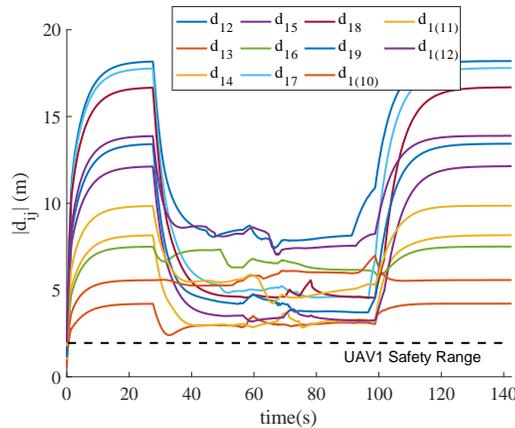

**Fig. 11.** Obtained results for the relative distance with 3D maneuvers

Fig. 11 shows the obtained relative distance between UAV1 with all other UAVs in the flock, while the horizontal dashed line in the figure shows the boundary of the safety range for the target UAV4. The same as Fig. 9, here, the solid lines plotted in the figure reveal the variation of relative distance $|d_{ij}|$ where, $\{(i,j)|i \neq j, j = 1,2,...,12\}$ between the UAVs and the target UAV4. It is obviously visible that all the neighboring UAVs have effectively utilized the inter-vehicle collision avoidance even during the formation change, performing obstacle avoidance maneuvers and formation recovery at the end. Note that, here we just plotted the results for the $i = 1$. On the other hand, results obtained in this section demonstrated the scalability of the proposed method in the previous section with 3D maneuvers without any complexity; the number of UAVs has increased from 8 to 12. Thus, for larger spaces in which larger barriers can be considered, this approach would be a good choice for the utilization of any number of UAVs for performing safe missions. (Notice: The related video for this scenario has been provided in the link available in the supplementary materials)

## IV. CONCLUSION

In the presented research, the demands for the applications of multi-UAV systems in urban areas in the presence of static and dynamic obstacles have been investigated. Then, inspired by the tilapia fish territorial behavior and self-organized obstacle avoidance of the pigeons, utilizing the probabilistic Lloyd's algorithm for centroidal Voronoi tessellation (CVT), and the rotation matrices, a semi-distributed nature-inspired collision-free formation control with a novel 3D obstacle detection and avoidance maneuvers for multi-UAV missions has been proposed. Further, to assess the validity of the proposed controller, implementing the planar maneuvers, initially it has been applied to a multi-UAV system control problem consisting of 8 UAVs in the presence of static and dynamic obstacles. Finally, utilizing the proposed novel 3D obstacle detection and avoidance, the proposed method has been applied to a larger scale flock of UAVs consisting of 12 vehicles in the presence of static and dynamic obstacles without any constraint on the method or the movement of obstacles. Results demonstrated the validity and the acceptable performance of the method in terms of formation change, performing the avoidance maneuvers, and formation recovery. For future works, the framework presented here can be extended to a learning-based neuromorphic digital-twin network for a more energy-efficient strategy.

## CONFLICT OF INTEREST

The authors declare that they have no conflicts of interest related to this research. The study was conducted in an objective and unbiased manner, and the results presented herein are based on rigorous analysis and interpretation. The authors have no financial or personal relationships with individuals or organizations that could potentially bias the findings or influence the conclusions of this study.## DATA AVAILABILITY

The data that support the findings of this study are openly available at https://github.com/INQUIRELAB/Multi-Agent-Drone-Control. All data were collected in accordance with the methods described in this paper.